# Performance Evaluation of Real-Time Object Detection for Electric Scooters

**Dong Chen, Ph.D.,[1] Arman Hosseini, M.S.,[2] Arik Smith, B.S.,[2] Amir Farzin Nikkhah, M.S.,[2] Arsalan Heydarian, Ph.D.,[3] Omid Shoghli, Ph.D.,[4] and Bradford Campbell, Ph.D.[5]**

[1]Environmental Institute & Link Lab & Computer Science, University of Virginia, Charlottesville, VA, 22903, USA; e-mail: dqc4vv@virginia.edu

[2]System Engineering, University of Virginia, Charlottesville, VA, 22903, USA; e-mail: qgn5zs@virginia.edu, ufy5tc@virginia.edu, kgc2mj@virginia.edu

[3]Link Lab & Engineering Systems and Environment, University of Virginia, Charlottesville, VA, 22903, USA; e-mail: heydarian@virginia.edu

[4]Civil Engineering Technology and Construction Management, University of North Carolina at Charlotte, Charlotte, NC 28223, USA; e-mail: oshoghli@charlotte.edu

[5]Link Lab & Computer Science, University of Virginia, Charlottesville, VA, 22903, USA; e-mail: bradjc@virginia.edu

**ABSTRACT**

Electric scooters (e-scooters) have rapidly emerged as a popular mode of transportation in urban areas, yet they pose significant safety challenges. In the United States, the rise of e-scooters has been marked by a concerning increase in related injuries and fatalities. Recently, while deep-learning object detection holds paramount significance in autonomous vehicles to avoid potential collisions, its application in the context of e-scooters remains relatively unexplored. This paper addresses this gap by assessing the effectiveness and efficiency of cutting-edge object detectors designed for e-scooters. To achieve this, the *first* comprehensive benchmark involving 22 state-of-the-art YOLO object detectors, including five versions (YOLOv3, YOLOv5, YOLOv6, YOLOv7, and YOLOv8), has been established for real-time traffic object detection using a self-collected dataset featuring e-scooters. The detection accuracy, measured in terms of mAP@0.5, ranges from 27.4% (YOLOv7-E6E) to 86.8% (YOLOv5s). All YOLO models, particularly YOLOv3-tiny, have displayed promising potential for real-time object detection in the context of e-scooters. Both the traffic scene dataset (https://zenodo.org/records/10578641) and software program codes (https://github.com/DongChen06/ScooterDet) for model benchmarking in this study are publicly available, which will not only improve e-scooter safety with advanced object detection but also lay the groundwork for tailored solutions, promising a safer and more sustainable urban micromobility landscape.

## INTRODUCTION

The integration of emerging modes of transportation, such as e-scooters, presents substantial safety challenges that hinder their smooth integration into our urban ecosystems. The emergence of e-scooters has been accompanied by an alarming surge in associated injuries and fatalities in the United States. Between 2017 and 2021, injuries associated with micromobility vehicles surged by 127% to 77,200. During this time, e-scooters witnessed the most significant rise in injuries and fatalities ([Tark, 2022](#)).

In the realm of autonomous vehicles, object detection algorithms have become immensely significant for ensuring safe driving. Specifically, deep learning-based algorithms have emerged as pivotal tools due to their capability to achieve high detection accuracy while demanding fewer computing resources, thereby becoming indispensable in autonomous driving systems ([Cai et al., 2021](#)). For example, [Cai et al. (2021)](#) highlight the development of a real-time object detector based on YOLOv4, emphasizing the paramount importance of obtaining both high accuracy and rapid inference speed. This breakthrough not only surpasses the accuracy of previous methods but also excels in real-time detection capabilities. Similarly, [Jia et al. (2023)](#) enhance accuracy and real-time performance in unmanned driving object detection with an improved YOLOv5. Despite the advancements in autonomous driving, the utilization of real-time object detection within the context of e-scooters remains relatively unexplored ([Apurv et al., 2021](#)). Object detection for e-scooters, as one of the fundamental computer vision tasks, can provide valuable and vital information to understand the surrounding environments for e-scooter users and facilitate the development of safety enhancement mechanisms.

Implementing current object detection algorithms on e-scooters presents challenges owing to various factors ([Alai et al., 2023](#)). First, the frequent vibrations generated by e-scooter movements disrupt sensor data, posing difficulties in real-time motion artifact mitigation. Second, e-scooters operate in diverse environments with varying lighting conditions, weather, and road surfaces, making it challenging for object detectors to adapt consistently. Additionally, reconciling algorithmic demands with the limited capacity of e-scooters introduces further complexity. Therefore, a crucial consideration lies in striking a balance between the effectiveness and efficiency of object detectors.

There are two primary types of deep learning object detectors: two-stage detectors, which involve a preprocessing step for object proposal generation in the initial stage, followed by object classification and bounding box regression in the subsequent stage. On the other hand, single/one-stage detectors are end-to-end, eliminating the necessity for the region proposal process. Compared to two-stage detectors like Faster-RCNN ([Ren et al., 2015](#)) and Mask-RCNN ([He et al., 2017](#)), one-stage detectors are more computationally efficient, faster in inference, and particularly suitable for real-time applications, especially on resource-constrained embedded devices like e-scooters. A notable example of a one-stage detector is You Only Look Once (YOLO), originally developed by [Redmon et al. (2016)](#) and further developed into YOLOv3 ([Redmon and Farhadi, 2018](#)). YOLOv3 strikes a balance between accuracy and speed, making it one of the most widely used object detectors. Following YOLOv3, architectural modifications have been introduced to enhance accuracy and/or speed, resulting in versions like YOLOv4 (Bochkovskiy et al., 2020), YOLOv5 (Jocher et al., 2020), YOLOv6 (Li et al., 2022a), YOLOv7 (Wang et al., 2022), and the latest YOLOv8 (Ultralytics 2023). These YOLO-derived object detectors can be configured with varying levels of model complexity, leading to different implementation variants.

As machine vision-based safety enhancement systems are evolving towards next-generation micro-mobility, differentiation among traffic objects and detection of individual traffic instances are becoming an increasingly important task. Research on multi-class traffic object detection/localization for e-scooters has been scant in literature (Cai et al., 2021). This is partly due to the lack of suitable traffic object datasets with multi-class bounding box annotations. Therefore, this study represents a unique contribution to the research community on traffic object detection for e-scooters by creating and releasing a 11-class traffic object dataset with over ten thousand of bounding box annotations collected under campus and urban traffic scenarios. Moreover, a comprehensive benchmark of state-of-the-art YOLO detectors, encompassing five versions (YOLOv3, YOLOv5, YOLOv6, YOLOv7, and YOLOv8), for object detection is built with impressive performance in terms of high detection accuracy and fast inference times. It is important to note that YOLOv4 is excluded from this study due to the unavailability of pre-trained models. Both the traffic scene dataset (https://zenodo.org/records/10578641) and software program codes (https://github.com/DongChen06/ScooterDet) for model benchmarking in this study are publicly available. This research is expected to have a significant impact on future studies aimed at developing machine vision-based object detection systems for enhancing e-scooter safety.

**Methods**

### Image Dataset

The traffic scene dataset used in this study, ScooterDet, is collected with Tobii Pro Glasses 3 and a Segway NineBot scooter. The collection route spanned from the University of Virginia campus to the urban area of Charlottesville, VA, USA. To ensure a diverse set of images for robust model performance, the dataset is collected mostly during the daytime. Image frames are extracted from the recorded videos taken by the Tobii Pro Glasses. Subsequently, data cleaning procedures are applied to remove low-quality images and those lacking relevant objects of interest.

The meticulously curated dataset is then labeled by trained personnel with bounding boxes around traffic objects in the images using the LabelMe tool (https://github.com/labelmeai/labelme). This study focuses on 11 specific objects: i.e., "person", "bicycle", "car", "truck", "bus", "traffic light", "fire hydrant", "stop sign", "bench", and "scooter". The resulting annotations, stored in JavaScript Object Notation (JSON) file format, then undergo visualization and double-checking by experts to ensure annotation accuracy and quality. This meticulous process results in a dataset comprising 2013 images covering 11 traffic object classes, with a total of 11,011 bounding box annotations, publicly accessible in the Zenodo repository (https://zenodo.org/records/10578641). This dataset will be continually updated with additional traffic scene images collected and labeled for future experiments.

### Experimentation

In this study, five versions of YOLO object detectors—YOLOv3, YOLOv5, YOLOv6, YOLOv7, and YOLOv8—are selected to develop traffic object detection models for the e-scooter dataset. Fig. 1 illustrates the modeling pipeline for traffic object detection, starting from data preparation to model training. The conversion process from the original image annotations in JSON format (from LabelMe) to YOLO format labels is a crucial initial step, ensuring compatibility with the YOLO training framework. After converting the formats, the dataset is then randomly divided into training, validation, and test subsets. This division follows a partition ratio of 60%, 20%, and 20%, corresponding to 1207, 402, and 404 images for each subset, respectively.

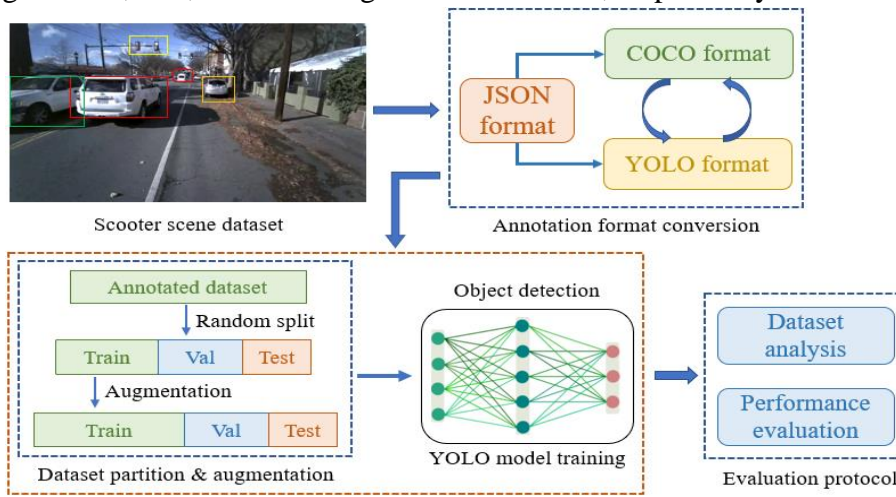

**Fig. 1. The proposed pipeline of object detection by YOLO object detectors.**

To enhance the training process, all YOLO object detection models undergo a phase of transfer learning, as described by Weiss et al. (2016), which involves refining the pre-existing weights acquired from training on the COCO dataset, as detailed by Lin et al. (2014). The original images are adjusted to a uniform size of $640 \times 640$ pixels to meet the input requirements of the YOLO architecture. These models are then trained using a batch size of 16 for 100 epochs within the PyTorch framework (version 2.1.2) (Paszke et al., 2019). To optimize the learning rate throughout the training period, a cosine annealing strategy is applied across all YOLO models, a method proposed by He et al. (2019). The entire process of model training and evaluation is carried out on a server running Ubuntu 20.04, which is configured with an AMD 7502 32-core processor (128 GB of RAM) and dual GeForce RTX 3090Ti GPUs (each with 24 GB of GDDR6X RAM).

### Performance Evaluation Metrics

The assessment of YOLO object detectors for identifying e-scooters involved evaluating detection accuracy, inference speeds, and model complexity. The quantity of model parameters indicates the complexity of the model, an important aspect for deploying models in practical settings. Generally, models with more parameters require additional memory for deployment and affect both computational costs and inference speeds. Detection accuracy, a crucial metric for object detection highlighted by Padilla et al. (2020), encompasses precision (P), recall (R), and mean average precision (mAP, notably mAP@0.5), with mAP being the key indicator for gauging the performance of object detectors in multi-class scenarios. Computational cost and inference time are assessed by metrics like floating point operations (FLOPs), which quantify the computational effort required to process a single instance, with FLOPs calculation facilitated by the THOP library. Inference time, the time it takes for a model to predict outcomes on an input image, is crucial for applications demanding real-time processing. This time was measured as the mean duration needed to analyze all images in the test dataset.

**RESULTS AND DISCUSSION**

### Performance of YOLO models

Table 1 summarizes the performance of various YOLO models on a test dataset for object detection, emphasizing their effectiveness in identifying 11 classes of traffic-related objects. Notably, the table includes different versions of YOLOv3—such as YOLOv3-tiny and YOLOv3, which incorporate feature pyramid networks (FPN), and YOLOv3-SPP, which utilizes spatial pyramid pooling (SPP) (Redmon and Farhadi, 2018). These models display remarkable accuracy levels, with mean average precision (mAP) at a threshold of 0.5 ranging from 27.4% for YOLOv7-E6E to a high of 86.8% for YOLOv5s. While most models (excluding YOLOv7-W6, YOLOv7-E6, YOLOv7-D6, and YOLOv7-E6E, which fall below 70%) achieve mAP@0.5 accuracies between 72.1% and 86.8%, with six models surpassing 85%. The lower performance of YOLOv7 variants is attributed to overfitting, suggesting the need for further exploration of solutions like data augmentation and generation techniques to enhance future model performance. The table also reveals that the majority of YOLO models tend to have higher precision than recall, indicating effective object detection capabilities. However, challenges remain in detecting smaller objects at a distance from the e-scooter, leading to missed detections and lower recall rates.

Model complexity and inference times are important for real-world application, especially in scenarios with limited resources like e-scooters. Fig. 2 displays the correlation between GFLOPs and inference times against the total number of parameters across all evaluated YOLO detectors, revealing a linear augmentation in GFLOPs and inference times as model parameters increase. YOLOv8x stands out for having the highest GFLOPs and the longest inference time at 29.5 milliseconds. Conversely, YOLOv5 variants—specifically YOLOv5n and YOLOv5s—and YOLOv3-tiny are highlighted for their superior computational efficiency and swift inference times (under 5 milliseconds). Additionally, Fig. 2 also illustrates model inference times versus mAP@0.5, indicating potential compromises in choosing models based on accuracy versus inference speed, where increased accuracy is often linked with longer inference durations. Despite these variances, all tested YOLO detectors are capable of real-time object detection, achieving processing rates of dozens or even hundreds of frames per second. Notably, YOLOv5 and YOLOv6 models exemplify an optimal balance between accuracy and efficiency. It is crucial to acknowledge these assessments are performed on high-end computing setups, with the performance on embedded systems yet to be determined.

**Table 1. Object detection performance of 22 YOLO detectors on the testing dataset.**

| Index | YOLO models | | Precision | Recall | mAP@0.5 |
|---|---|---|---|---|---|
| 1 | YOLOv3 | YOOv3-tiny | 0.747 | 0.680 | 0.721 |
| 2 | | YOLOv3 | 0.854 | 0.842 | 0.857 |
| 3 | | YOLOv3-SPP | 0.841 | 0.844 | 0.855 |
| | | *Average* | *0.814* | *0.789* | *0.811* |
| 4 | YOLOv5 | YOLOv5n | 0.673 | 0.789 | 0.797 |
| 5 | | YOLOv5s | **0.912** | 0.812 | **0.868** |
| 6 | | YOLOv5m | 0.855 | 0.812 | 0.849 |
| 7 | | YOLOv5l | 0.871 | 0.850 | 0.866 |
| 8 | | YOLOv5x | 0.826 | 0.841 | 0.846 |
| | | *Average* | *0.827* | *0.821* | *0.845* |
| 9 | YOLOv6 | YOLOv6n | 0.832 | 0.821 | 0.841 |
| 10 | | YOLOv6s | 0.822 | 0.814 | 0.841 |
| 11 | | YOLOv6m | 0.842 | 0.833 | 0.857 |
| | | *Average* | *0.832* | *0.823* | *0.846* |
| 12 | YOLOv7 | YOLOv7 | 0.830 | 0.802 | 0.808 |
| 13 | | YOLOv7x | 0.857 | **0.876** | 0.862 |
| 14 | | YOLOv7-W6 | 0.705 | 0.517 | 0.583 |
| 15 | | YOLOv7-E6 | 0.680 | 0.543 | 0.601 |
| 16 | | YOLOv7-D6 | 0.478 | 0.516 | 0.470 |
| 17 | | YOLOv7-E6E | 0.561 | 0.241 | 0.274 |
| | | *Average* | *0.509* | *0.491* | *0.457* |
| 18 | YOLOv8 | YOLOv8n | 0.838 | 0.736 | 0.804 |
| 19 | | YOLOv8s | 0.841 | 0.784 | 0.818 |
| 20 | | YOLOv8m | 0.842 | 0.767 | 0.795 |
| 21 | | YOLOv8l | 0.869 | 0.752 | 0.797 |
| 22 | | YOLOv8x | 0.843 | 0.732 | 0.809 |
| | | *Average* | *0.847* | *0.754* | *0.805* |

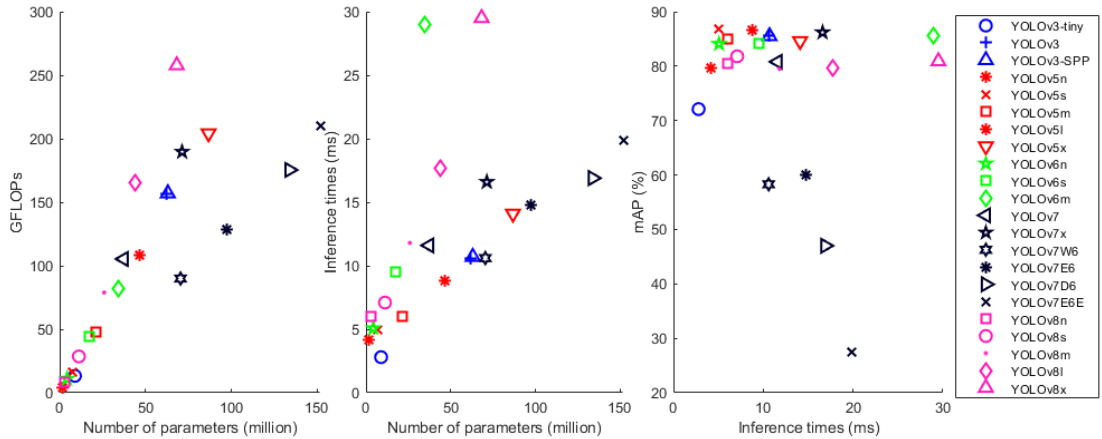

**Fig. 2. GFLOPs and inference time v.s. number of parameters (million), and mAP@0.5 versus inference times. The detection models within the same type of YOLO detectors are labeled with markers of the same color. GFLOPs stands for giga float point operations (FPLOPs), which is equal to 109 FLOPs, and mAP denotes mean average precision.**

Figure 3 presents illustrative examples of image predictions made by YOLOv5s, showing the model's capability to generate visually accurate predictions across a range of scenarios, including those with diverse and cluttered backgrounds typical of densely populated urban environments. The predictive performance of YOLOv5s on test images has been further compiled into video formats for enhanced illustration, with these videos being made available on our GitHub page (https://github.com/DongChen06/ScooterDet).

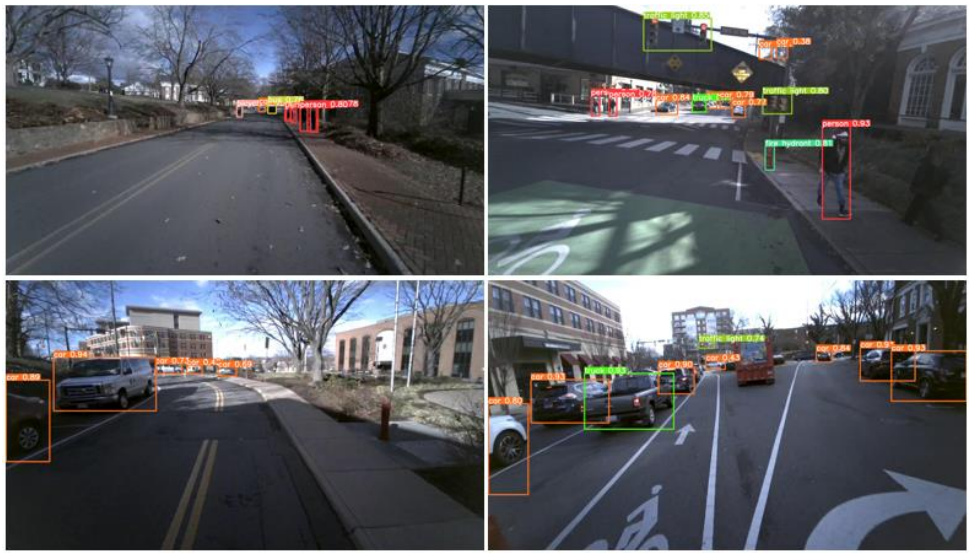

**Fig. 3. Examples of traffic scene images with predicted bounding boxes.**

These outcomes collectively highlight the effectiveness of the chosen YOLO object detectors in accurately identifying multiple classes of traffic objects in the context of e-scooter use. The ability of these detectors to maintain high accuracy in complex urban settings underscores their potential utility in enhancing the safety and navigational efficacy of e-scooters, further demonstrating their contribution to the field of machine vision-based safety systems for micro-mobility solutions.

### Class-wise performance of selected YOLO models

In this subsection, in-depth analysis of detection accuracies for specific traffic object classes is conducted, focusing solely on YOLOv5s and YOLOv8s due to limitations in space, as detailed in Table 2. The accuracy in detecting traffic objects can be influenced by various factors, including the quantity and dimensions of bounding boxes, class variability, and similarities among object classes. Within the testing dataset, which included 2,160 instances, the "car" category has the most annotations (1,468) and shows high mAP@0.5, achieving 92.4% for YOLOv5s and 92.1% for YOLOv8s. Other categories such as "bicycle", "motorcycle", "bus", "fire hydrant", and "stop sign" also yield mAP@0.5 scores above 90%. Despite having nearly 200 bounding box annotations, the "person" class is detected with comparatively lower accuracy by both YOLOv5s and YOLOv8s, with mAP@0.5 scores of 79.7% and 77.2%, respectively. This lower accuracy is likely due to the small and blurred appearances of persons in the images, complicating accurate localization. Furthermore, the "bench" and "scooter" classes are particularly challenging for both detectors, with mAP scores falling below 80%. The limited number of annotations for these classes may contribute to their lower detection accuracies. To improve detection accuracy for these more challenging classes, incorporating a greater variety of training samples and refining training methodologies may be beneficial.

**Table 2. Class-wise performance of YOLOv5s and YOLOv8s. P, R and mAP represent precision, recall and mean average precision, respectively.**

| Index | Object Class | #instances | YOLOv5s | | | YOLOv8s | | |
|---|---|---|---|---|---|---|---|---|
| | | | P | R | mAP@0.5 | P | R | mAP@0.5 |
| 1 | person | 224 | 0.876 | 0.694 | 0.797 | 0.821 | 0.772 | 0.772 |
| 2 | bicycle | 14 | 0.907 | 0.929 | 0.946 | 0.779 | 0.766 | 0.766 |
| 3 | car | 1468 | 0.920 | 0.866 | 0.924 | 0.870 | 0.921 | 0.921 |
| 4 | motorcycle | 11 | 0.867 | 0.818 | 0.906 | 0.842 | 0.965 | 0.965 |
| 5 | bus | 34 | 0.977 | 0.971 | 0.989 | 0.949 | 0.963 | 0.963 |
| 6 | truck | 146 | 0.869 | 0.864 | 0.895 | 0.873 | 0.887 | 0.887 |
| 7 | traffic light | 184 | 0.935 | 0.783 | 0.859 | 0.864 | 0.817 | 0.817 |
| 8 | fire hydrant | 45 | 0.918 | 0.844 | 0.902 | 0.813 | 0.814 | 0.814 |
| 9 | stop sign | 11 | 1.00 | 0.790 | 0.950 | 0.901 | 0.905 | 0.905 |
| 10 | bench | 18 | 0.840 | 0.778 | 0.776 | 0.759 | 0.692 | 0.692 |
| 11 | scooter | 5 | 0.926 | 0.600 | 0.604 | 0.775 | 0.493 | 0.493 |
| | *All* | *2160* | *0.912* | *0.812* | *0.868* | *0.841* | *0.784* | *0.818* |

**CONCLUSION**

Real-time object detection plays a crucial role in enhancing e-scooter safety, with a key consideration being the balance between the effectiveness and efficiency of object detectors. This balance is especially vital for real-time implementations in e-scooters, given the constraints of limited on-board computing resources. This paper presented, to date, a comprehensive performance evaluation on state-of-the-art object detectors for e-scooters. A comprehensive benchmark suite of 22 selected YOLO object detectors was established for object detection for e-scooters, which was evaluated in terms of detection accuracies, model complexity, and inference time. The YOLO detectors achieved mAP@0.5 from 0.274 by YOLOv7-E6E to 0.868 by YOLOv5s. Overall, YOLOv5 yielded the best trade-off between efficiency and effectiveness. This performance benchmark is expected to serve as a reference source for informed choices of YOLO object detectors in the context of e-scooters for object detection and can also be beneficial for object detection tasks in urban micro-mobility.